\documentclass[runningheads]{llncs}

\usepackage{eccv}

\usepackage{eccvabbrv}

\usepackage{graphicx}
\usepackage{booktabs}
\usepackage{siunitx}
\usepackage{multirow}
\usepackage{wrapfig}

\usepackage{pgfplots}
\usepackage{pgfplotstable}
\pgfplotsset{compat=1.18}

\usepackage[accsupp]{
  axessibility
} 

\usepackage{hyperref}

\usepackage{orcidlink}

\usepackage{array}
\newcommand*{\rot}{\rotatebox{90}}

\begin{document}

  \title{UltraViT: Latency-Optimized On-device Vision Encoder for Large Vision-Language
  Models}

  \titlerunning{UltraViT: Latency-Optimized On-device Vision Encoder}

  \author{Ioannis Maniadis Metaxas\thanks{Equal contribution}\inst{1}\orcidlink{0009-0000-9588-6913} \and 
  Adrian Bulat$^\star$\inst{1,2}\orcidlink{0000-0002-3185-4979} \and 
  Alberto Baldrati$^\star$\inst{1}\orcidlink{0000-0002-5012-5800} \and 
  Anestis Zaganidis\inst{1}\orcidlink{0000-0002-1198-2201} \and 
  Yassine Ouali\inst{1}\orcidlink{0000-0001-9227-4487} \and 
  Hyeonuk Kim\inst{1}\orcidlink{0000-0001-5216-2552} \and 
  Georgios Tzimiropoulos\inst{1,3}\orcidlink{0000-0002-1803-5338}
  }

  \authorrunning{I. Maniadis Metaxas et al.}

  \institute{Samsung AI Cambridge \and 
  Technical University of Iasi \and 
  Queen Mary University of London}

  \maketitle

  \begin{abstract}
    Large Vision-Language Models (LVLMs) remain bottlenecked by massive computational
    footprints, precluding their deployment on resource-constrained edge devices.
    While efforts to compress LVLMs focus heavily on vision token reduction or smaller
    language models, the vision encoder is largely overlooked, typically deployed
    as a monolithic, computationally heavy feature extractor. Moreover, there is
    no previous effort that designs a vision encoder for LVLMs directly
    optimized for on-device latency. In this paper, we present UltraViT, a vision
    encoder for LVLMs, explicitly designed and optimized for on-device
    performance. Specifically, by taking into account real on-device latencies, we
    systematically design a pyramidal architecture that strategically integrates
    and adapts heterogeneous spatial mixers at the macro-block level.
    Furthermore, to pre-train UltraViT, we propose a novel two-stage generative
    pre-training strategy: cultivating rich spatial features via dense distillation,
    followed by direct generative supervision from a capacity-mixed frozen LLM. Compared
    to standard contrastive and SSL, we show that our pre-training is much more
    effective for achieving high-level semantic grounding for UltraViT needed
    for the subsequent generative multimodal alignment of LVLM training. Extensive
    experiments demonstrate that our on-device latency-informed design combined with
    our tailored training strategy establishes a new state-of-the-art for efficient
    LVLM encoding, significantly outperforming existing encoder-centric
    baselines while operating on-device at nearly $1.7 \times$ the speed.

  \end{abstract}

  \section{Introduction}
  \label{sec:intro}

  The recent proliferation of Large Vision-Language Models (LVLMs)~\cite{wang2024qwen2vl,li2024llava-ov,zhu2025internvl3}
  has driven unprecedented advancements across a broad spectrum of multimodal
  tasks~\cite{zhang2021mme,liu2024mmbench,kembhavi2016diagram,liu2024ocrbench,mathew2022infographicvqa,methani2020plotqa}.
  However, these models are characterized by massive parameter counts and
  immense computational footprints, fundamentally hindering their deployment on resource-constrained
  mobile and edge devices. To mitigate these high inference costs, prior efforts
  targeting efficient LVLMs have predominantly focused on two strategies:
  aggressively reducing the length of the visual context via token sparsification or
  compression~\cite{zhang2024sparsevlm,bulat2025compress,yang2025visionzip,cai2024matryoshka},
  and/or coupling lighter language models with more computationally efficient cross-modal
  interaction mechanisms~\cite{chu2024mobilevlm,marafioti2025smolvlm}. However, with
  a few exceptions (i.e., FastVLM~\cite{vasu2025fastvlm}), existing literature largely
  overlooks the vision encoder architecture itself, despite the fact that
  relying on heavyweight models like CLIP-ViT-L~\cite{radford2021learning} or
  SigLIP-400M~\cite{zhai2023sigmoid} as fixed feature extractors creates a
  severe computational bottleneck, especially for the processing of higher-resolution
  images and video.

  Concurrently, while significant progress has been made in designing efficient,
  mobile-friendly Vision Transformers (ViTs)~\cite{pan2022edgevits,vasu2023fastvit,yun2024shvit},
  these architectures have been heavily optimized for the task of ImageNet classification,
  rather than the dense image understanding and reasoning demanded by LVLMs. Crucially,
  prior efficient ViT paradigms typically introduce a specific spatial mixer (e.g.,
  sparse attention~\cite{pan2022edgevits} or depthwise convolutional modules~\cite{sandler2018mobilenetv2}), and replicate
  it uniformly across the entire network or combined with vanilla attention. We
  argue that this homogenous block formulation is suboptimal for \textit{on-device}
  LVLM image encoding when latency considerations must be taken into account.

  To bridge this gap, we present UltraViT, a highly efficient, pyramidal vision architecture
  explicitly optimized for on-device LVLM deployment. Rather than relying on a
  monolithic block design, we systematically engineer a heterogeneous architecture
  by strategically selecting distinct spatial mixers across different network stages,
  and further adapting their underlying operations specifically for NPU/mobile
  execution. UltraViT is designed to be latency-efficient for on-device deployment (measured on a real edge device) and to possess dense representation capabilities necessary for multimodal
  reasoning. Moreover, by designing the vision encoder from the ground-up to be inherently
  compact and efficient, we entirely circumvent the need for post-hoc vision
  token compressors.

  Beyond architectural innovation, we propose a highly effective pre-training strategy for UltraViT. Our motivation is as follows: First, standard contrastive training (e.g., CLIP~\cite{radford2021learning}) represents too coarse a signal for LVLMs, and is misaligned with generative token generation \footnote{This misalignment is consistently evidenced by the
  empirical necessity of discarding the final transformer layer before LLM
  integration~\cite{bolya2025perception}.}. Furthermore, although Masked Image Modeling (MIM)
  and DINO~\cite{oquab2023dinov2} objectives yield
  strong dense localization, their reconstructive nature produces low-level visual
  features that lack the high-level semantics required by an LLM. To alleviate this, we introduce a novel, two-stage pre-training paradigm. First, we
  perform dense feature distillation from a robust, high-resolution teacher model
  to cultivate rich spatial representations. Subsequently, we attach the encoder
  to a frozen vision-enabled LLM to provide a direct generative training signal.
  To strictly maintain computational efficiency during this second phase, we devise
  a dynamic LLM mixing strategy that samples and mixes LLMs of varying capacities.

  In summary, our combined architectural optimization and generative pre-training enables UltraViT to establish a new state-of-the-art
  for on-device LVLMs. Our main contributions are summarized as follows:
  \begin{itemize}
    \item We introduce UltraViT, a latency-efficient pyramidal vision encoder
    explicitly tailored for on-device deployment of LVLMs. Informed by real latency measurements, and by systematically exploring heterogeneous spatial mixers, we overcome the limitations of previously proposed homogeneous ViT architectures for dense semantic reasoning.

    \item We propose a novel two-stage generative pre-training paradigm tailored to UltraViT. Our approach circumvents fundamental misalignments of contrastive pre-training by leveraging dense spatial distillation followed by direct generative supervision from a frozen, capacity-mixed LLM.

    \item Extensive experiments demonstrate that UltraViT achieves state-of-the-art on-device performance, outperforming prior encoder-centric computational baselines, such as FastVLM~\cite{vasu2025fastvlm}, while being almost $1.7\times$ faster.
  \end{itemize}

  \section{Related Work}

  \noindent
  \textbf{Efficient Vision Encoders:} Constructing efficient vision architectures
  is a long-standing challenge in computer vision. Based on their underlying
  operator primitives, existing models can be broadly classified into three categories:
  convolutional, transformer-based, and hybrid architectures.

  Historically, early architectures were predominantly driven by Convolutional Neural
  Networks (CNNs). Research in this direction yielded seminal efficiency
  techniques such as depth-wise convolutions and inverted residuals in
  MobileNets~\cite{sandler2018mobilenetv2,howard2019searching}, channel shuffling
  in ShuffleNets~\cite{zhang2018shufflenet,ma2018shufflenet}, cheap linear
  transformations in GhostNet~\cite{han2020ghostnet}, compound scaling laws in EfficientNet~\cite{tan2019efficientnet},
  and structural re-parameterization~\cite{ding2021repvgg,vasu2023fastvit}. With
  the advent of the Vision Transformer~\cite{dosovitskiy2020image}, which
  demonstrated superior scaling properties compared to CNNs, numerous works have
  attempted to alleviate the quadratic complexity of global self-attention.
  Prevalent strategies include adopting pyramidal structural topologies~\cite{wang2021pyramid},
  restricting attention to local windows~\cite{liu2021swin}, and utilizing
  efficient primitives such as separable self-attention~\cite{mehta2022separable},
  sparse attention~\cite{pan2022edgevits}, or single-head partial attention~\cite{yun2024shvit}.
  Current state-of-the-art vision encoders commonly combine these two paradigms,
  utilizing convolutional blocks in the earlier, high-resolution stages to capture
  local features, and attentive blocks in the later stages to model global
  dependencies~\cite{pan2022edgevits,vasu2023fastvit,yun2024shvit}. However,
  except for FastVLM~\cite{vasu2025fastvlm} none of these
  architectures are designed with LVLMs in mind. In contrast to the aforementioned
  works, we design our approach from the ground up with LVLMs in mind, carefully
  adapting and then selecting in a principled manner appropriate mixers depending
  on the location within the model, constructing a novel heterogeneous
  architecture.

  \noindent
  \textbf{Efficient Vision-Language Models:} Despite their unrivaled accuracy and
  reasoning capabilities, LVLMs are often computationally prohibitive for on-device
  deployment. Current research aimed at mitigating this computational bottleneck
  generally follows two main trajectories: (1) reducing the sequence length of
  visual tokens, either via training-aware compression modules~\cite{chu2024mobilevlm,yang2025visionzip,cai2025matryoshka,hu2024matryoshka}
  or through post-hoc, zero-shot token pruning and attention sparsification
  techniques~\cite{zhang2024sparsevlm,zhang2024cls,xing2025pyramiddrop,arif2025hired},
  and (2) coupling the vision encoder with smaller, highly optimized language
  models~\cite{chu2024mobilevlm,deitke2025molmo,marafioti2025smolvlm}. While these
  methods yield favorable speedups across various accuracy-latency trade-offs,
  the vast majority continue to rely on heavy, computationally rigid vision encoders,
  typically the 400M parameter SigLIP backbone, not designed with LVLMs in mind.
  As the LLM backbones become increasingly compact (e.g., scaling down to 0.5B
  or 1.5B parameters for edge deployment), the relative latency burden of the vision
  encoder becomes disproportionately large. For instance, in an edge-oriented LLaVA-OV
  model comprising a SigLIP-400M vision encoder and a 1.5B Qwen2.5 language model, for a 1024x1024px image
  the visual feature extraction phase alone can account for nearly 50\% (!) of the
  total end-to-end inference time on mobile devices. Consequently, in such
  resource-constrained environments, optimizing the vision encoder itself,
  rather than exclusively attempting to shrink the LLM sequence length, represents
  a critical, yet under-explored avenue for reducing latency. Only
  very recently has FastVLM~\cite{vasu2025fastvlm} introduced a custom architecture
  specifically aimed at efficient LVLM encoding. In contrast to FastVLM, we introduce
  a novel heterogeneous architecture coupled with an efficient two-stage training
  paradigm. By combining dense distillation with LLM-in-the-loop generative
  pre-training, we deviate from traditional contrastive methods, establishing a
  new state-of-the-art that surpasses FastVLM in accuracy while operating
  significantly faster.

  \noindent
  \textbf{Vision Encoder pre-training for LVLMs:} The dominant paradigm for pre-training
  vision encoders heavily relies on contrastive learning objectives applied to massive
  image-text datasets, epitomized by CLIP~\cite{radford2021learning} and SigLIP~\cite{zhai2023sigmoid}.
  While dense, representation-focused objectives such as Masked Image Modeling (MIM)
  or DINO have been explored, they generally underperform their contrastive counterparts
  in LVLM settings~\cite{cocchi2025llava,liu2025data}. Similarly, directly applying
  captioning losses (e.g., SigLIP~2~\cite{tschannen2025siglip}) during vision encoder
  pre-training yields downstream performance comparable to variants trained without
  such losses (e.g., SigLIP~\cite{zhai2023sigmoid}), suggesting limited marginal
  utility in the presented form~\cite{cocchi2025llava}. Hence, current state-of-the-art
  is represented by contrastively trained model~\cite{tschannen2025siglip}, often
  combined with score distillation objectives~\cite{vasu2025fastvlm}.

  \section{Architecture optimization}

  \begin{figure}[!ht]
    \centering
    \includegraphics[width=1\linewidth]{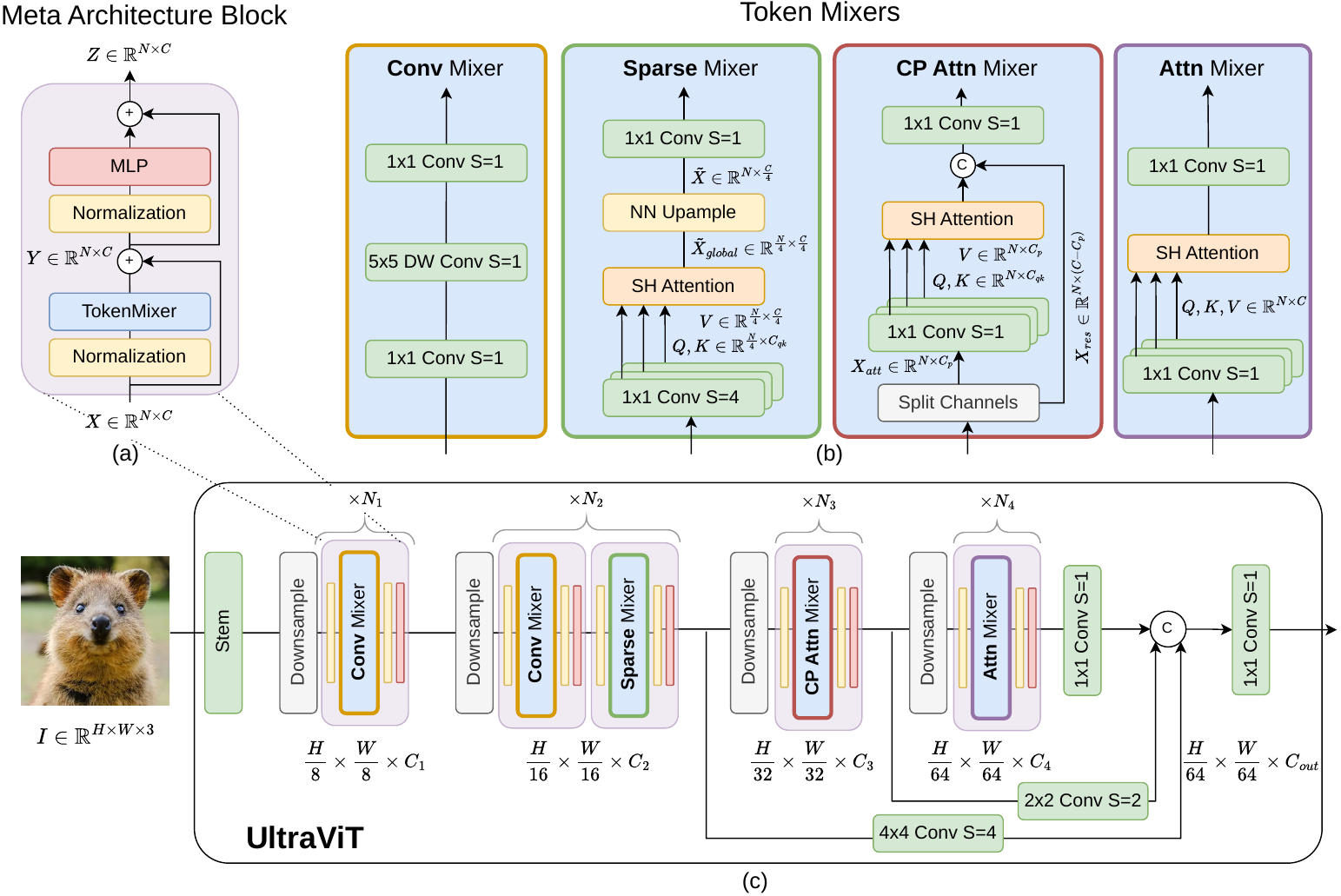}
    \caption{(a) General Structure of the Meta Architecture Block. (b): Architecture of the four different token mixers employed. (c) Overview of the full UltraViT architecture. }
    \label{fig:ultravit}
  \end{figure}

  At its core, an LVLM is composed of a vision encoder, a cross-modal alignment projector, and an LLM. Let $I$ denote a high-resolution
  input image. The vision encoder acts as the frontline feature extractor, digesting $I$ into a sequence of visual tokens
  $\mathbf{F}\in \mathbb{R}^{N \times C}$. A lightweight MLP projector then aligns these visual features with the language modality, allowing the LLM to auto-regressively condition on both the image and the query text. In this work, we adopt the widely-used LLaVA-OV~\cite{li2024llava-ov} design. 

  Unlike the bulk of prior work on efficient LVLMs, we focus on the design of the vision encoder specifically for efficient on-device deployment. Recognizing the limitations of rigid, homogenous ViT backbones, we pivot to a heterogeneous macro-search
  space. The proposed UltraViT adopts a classic four-stage pyramidal~\cite{wang2021pyramid} topology
  to progressively aggregate spatial information and reduce the token sequence length. However, unlike prior efficient models (e.g., EdgeViT~\cite{pan2022edgevits},
  SH-ViT~\cite{yun2024shvit}) that apply a single (or two) type of spatial mixers across all
  stages, we hypothesize that different depths of the network necessitate
  fundamentally different inductive biases and computational primitives. Consequently, we systematically design and adapt distinct token mixers, specifically targeting NPU and mobile execution constraints to achieve an optimal balance of latency and representational
  density.

  \subsection{Heterogeneous Spatial Mixers}
  \label{sec:spatial_mixers}

  Starting from the foundational transformer block structure of~\cite{vaswani2017attention}, recent studies~\cite{yu2022metaformer} have convincingly demonstrated that the general macro-structure of these blocks is the true catalyst for their success. Specifically, the arrangement of residual connections flanking a primary
  token mixer and a subsequent channel mixer (MLP) is paramount to the representation quality. Formally, let $\mathbf{X}\in \mathbb{R}^{N \times C}$ denote the input
  sequence of visual tokens, where $N = H \times W$ is the spatial resolution
  and $C$ is the embedding dimension. The general Meta Architecture Block is defined as:
  \begin{align}
    \mathbf{Y} & = \mathrm{TokenMixer}(\mathrm{Norm}(\mathbf{X})) + \mathbf{X}\label{eq:metaformer_1} \\
    \mathbf{Z} & = \mathrm{MLP}(\mathrm{Norm}(\mathbf{Y})) + \mathbf{Y}\label{eq:metaformer_2}
  \end{align}
  Building upon this insight, UltraViT strictly preserves this proven macro-architecture across all network depths: we standardize the block topology (Eqs.~\ref{eq:metaformer_1},~\ref{eq:metaformer_2})
  and modularly interchange the internal $\mathrm{TokenMixer}(\cdot)$ module.

  In the following, without loss of generality, we treat $\mathbf{X}$ interchangeably as
  $\mathbf{X} \in \mathbb{R}^{N \times C}$ and $\mathbf{X} \in \mathbb{R}^{H \times W \times C}$ ($N = H \times W$). The reshaping required to
  transition between grid-based operations (e.g., convolutions on $\mathbf{X} \in \mathbb{R}
  ^{H \times W \times C}$) and sequence-based operations (e.g., attention on $\mathbf{X} \in
  \mathbb{R}^{N \times C}$) is left implicit.

  To optimize for both on-device latency and dense representational capacity
  demanded by LVLMs, we consider, adapt, and evaluate four distinct classes of spatial mixers:

  \noindent
  \textbf{Convolutional Mixer:} Convolutional layers are a natural choice for spatial
  mixing due to their inductive biases for locality and translation equivariance.
  They are a prime candidate for the initial, high-resolution stages of the network,
  where the sequence length $N$ is excessively large, rendering global routing computationally
  prohibitive. Moreover, they are well-supported by mobile NPUs. Following prior
  work~\cite{trockman2022patches,vasu2023fastvit}, we use DepthWise (DWConv)
  and Pointwise Convolutions in the spatial mixer. The convolutional mixer is defined
  as:
  \begin{gather}
    \mathbf{X}_{proj}= \mathrm{Conv}_{1\times1}(\mathbf{X}) \\
    \mathbf{Y}= \mathrm{Conv}_{1\times1}(\mathrm{DWConv}_{5\times5}(\mathbf{X}_{proj}
    ))
  \end{gather}

  \noindent
  \textbf{Sparse Mixers:} While convolutions are efficient, they struggle to capture long-range dependencies that may be desirable in LVLMs. Conversely, standard self-attention
  is computationally expensive. As a middle ground, sparse attention mechanisms have
  been proposed~\cite{child2019generating,beltagy2020longformer,pan2022edgevits}.
  We use EdgeViT's~\cite{pan2022edgevits}
  sparse attention as a starting point:
  \begin{gather}
    \mathbf{X}_{sparse}= \mathcal{S}(\mathbf{X}) \\
    \mathbf{\Tilde{X}}_{global}= \mathrm{Attention}(\mathbf{X}_{sparse}W^{Q}, \mathbf{X}
    _{sparse}W^{K}, \mathbf{X}_{sparse}W^{V}) \\
    \mathbf{Y}= \mathrm{ConvTrans_{DW}}(\mathbf{\Tilde{X}}_{global}),
  \end{gather} where $\mathcal{S}$ denotes the sparse sampling operator.

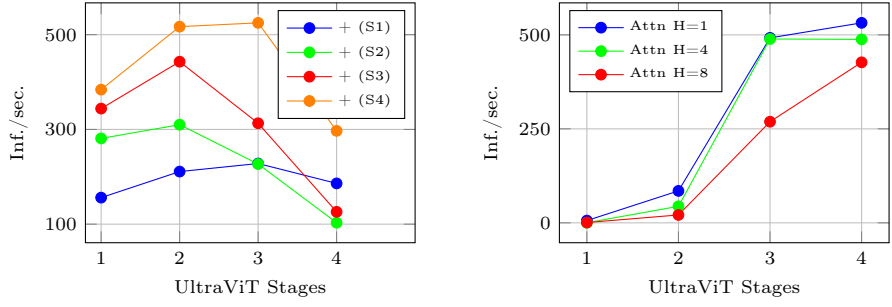
\begin{figure}[!ht]
  \centering
  \begin{subfigure}{\dimexpr\textwidth/2-0.5em\relax}
    \centering
    \begin{tikzpicture}
      \begin{axis}[
        xlabel={UltraViT Stages},
        ylabel={Inf./sec.},
        xmin=-1,
        xmax=3.2,
        legend style={font=\tiny},
        x dir=reverse,
        grid=major,
        xtick={0,1,2,3},
        xticklabels={4,3,2,1},
        ytick={100,300,500},
        legend pos=north east,
        font=\scriptsize,
        width=\linewidth,
        height=0.8\linewidth,
      ]
        \addplot[ blue, mark=*, table/skip first n=1 ] table [x index=0, y index=1, col sep=comma] {csvs/edgevit_blocks.csv}; \addlegendentry{+ (S1)}
        \addplot[ green, mark=*, table/skip first n=1 ] table [x index=0, y index=2, col sep=comma] {csvs/edgevit_blocks.csv}; \addlegendentry{+ (S2)} 
        \addplot[ red, mark=*, table/skip first n=1 ] table [x index=0, y index=3, col sep=comma] {csvs/edgevit_blocks.csv}; \addlegendentry{+ (S3)} 
        \addplot[ orange, mark=*, table/skip first n=1 ] table [x index=0, y index=4, col sep=comma] {csvs/edgevit_blocks.csv}; \addlegendentry{+ (S4)}
      \end{axis}
    \end{tikzpicture}
    \label{fig:edgevit}
  \end{subfigure}\hfill
  \begin{subfigure}{\dimexpr\textwidth/2-0.5em\relax}
    \centering
    \begin{tikzpicture}
      \begin{axis}[
        xlabel={UltraViT Stages},
        ylabel={Inf./sec.},
        legend style={font=\tiny},
        x dir=reverse,
        grid=major,
        xtick={0,1,2,3},
        xticklabels={4,3,2,1},
        ytick={0,250,500},
        legend pos=north west,
        font=\scriptsize,
        width=\linewidth,
        height=0.8\linewidth,
      ]
        \addplot[ blue, mark=*, table/skip first n=1 ] table [x index=0, y index=1, col sep=comma] {csvs/attn_heads.csv}; \addlegendentry{Attn H=1}
        \addplot[ green, mark=*, table/skip first n=1 ] table [x index=0, y index=2, col sep=comma] {csvs/attn_heads.csv}; \addlegendentry{Attn H=4} 
        \addplot[ red, mark=*, table/skip first n=1 ] table [x index=0, y index=3, col sep=comma] {csvs/attn_heads.csv}; \addlegendentry{Attn H=8}
      \end{axis}
    \end{tikzpicture}
    \label{fig:attn}
  \end{subfigure}
  \caption{Inferences per second across UltraVit stages. Left: the throughput when the sparse attention mixer is applied cumulatively up to the i-th stage.
  Right: stage-wise throughput with attention mixers with a varying number of attention heads.}
  \label{fig:sparse_single_inference}
\end{figure}

  While conceptually effective, this instantiation incurs notable
  latency penalties when deployed on mobile NPUs. To circumvent this, we propose a new optimized Sparse mixer.
  Specifically, we introduce four core adaptations: (S1)
  we replace multi-head attention with single-head
  attention; (S2) we fuse the discrete
  sampling operator $\mathcal{S}$ directly into the
  $\mathbf{Q}, \mathbf{K}, \mathbf{V}$ projections by implementing them as
  strided convolutions; (S3) we replace the transposed depthwise convolution
  ($\mathrm{ConvTrans_{DW}}$) with a hardware-friendly nearest-neighbor upsampling
   operation; and crucially, (S4) we reduce the dimensionality of the attention projections by mapping queries and keys to a lower-dimensional space and compressing the value projections by a factor of 4.
  Fig.~\ref{fig:sparse_single_inference} (left) shows the cumulative effect of applying these structural refinements, with each
  yielding substantial on-device speed-ups. The new mixer is defined as:
  \begin{gather}
    \mathbf{Q}, \mathbf{K}, \mathbf{V}= \mathrm{Conv_{stride}}(\mathbf{X}; W^{Q})
    , \, \mathrm{Conv_{stride}}(\mathbf{X}; W^{K}), \, \mathrm{Conv_{stride}}(\mathbf{X}
    ; W^{V}) \\
    \mathbf{\Tilde{X}}_{global}= \mathrm{SH \text{-}Attention}(\mathbf{Q}, \mathbf{K}
    , \mathbf{V}) \\
    \mathbf{Y}= \mathrm{Upsample_{NN}}(\mathbf{\Tilde{X}}_{global})W^{O},
  \end{gather} where $W^{Q}, W^{K}\in \mathbb{R}^{C \times C_{qk}}$ and $W^{V}\in \mathbb{R}^{C \times \frac{C}{4}}$
  specify the channel-reduced projection matrices, and $W^{O}\in \mathbb{R}^{\frac{C}{4}
  \times C}$ represents the output projection layer. In all our experiments, we set $C_{qk} = 16$.

  \noindent
  \textbf{Channel-Partitioned (CP) Attention Mixer:} Traditional Multi-Head Self
  Attention (MHSA) splits the feature dimension $C$ across multiple independent
  attention heads, causing memory fragmentation and bloated memory access patterns on edge devices. Drawing from SH-ViT~\cite{yun2024shvit}, we employ Channel-Partitioned
  Self-Attention (CPSA), which computes single-head attention over a subset of the embedding dimension:
  \begin{gather}
    \mathbf{X}_{att}, \mathbf{X}_{res}= \mathrm{Split}(\mathbf{X}, [C_{p}, C - C_{p}
    ]) \\
    \mathbf{\Tilde{X}}_{att}= \mathrm{SH\text{-}Attention}(\mathbf{X}_{att}W^{Q},
    \mathbf{X}_{att}W^{K}, \mathbf{X}_{att}W^{V}) \\
    \mathbf{Y}= \mathrm{Concat}(\mathbf{\Tilde{X}}_{att}, \mathbf{X}_{res})W^{O},
  \end{gather}
where $W^{Q}, W^{K}\in \mathbb{R}^{C \times C_{qk}}$ and $W^{V}\in \mathbb{R}^{C \times C_{p}}$ are the attention projection matrices, and $W^{O} \in \mathbb{R}^{C \times C}$ is the output projection matrix. $C_{p}$ and $C_{qk}$ denote the number of channels used for attention and the key-query dimension, respectively. Following SH-ViT~\cite{yun2024shvit}, we set $C_{qk}=16$ and $C_{p} = \frac{C}{4}$ in all experiments.

  \noindent
  \textbf{Vanilla Attention Mixer:} In the final stages where spatial downsampling reduces $N$ significantly, attention's complexity
  is no longer the primary bottleneck. Hence, vanilla attention can be beneficial thanks to its stronger representational capacity. However, we opt for a single-head attention mechanism instead of the traditional multi-head one. This design choice is tailored to on-device deployment: multi-head attention introduces
  significant memory access overhead due to the partitioning of the embedding dimension, which leads to fragmented tensor operations and sub-optimal utilization of mobile NPUs. Single-head attention, conversely, simplifies data access patterns and maintains a contiguous memory layout, enabling faster execution and lower latency on edge devices while still harnessing the global receptive field benefits of standard attention. Note that, unlike the CP Attention Mixer, the vanilla attention mixer updates all channels. Fig.~\ref{fig:sparse_single_inference} (right) shows the performance comparison between vanilla attention and single-head attention.
  \vspace{1cm}

  \subsection{Block Selection Process}

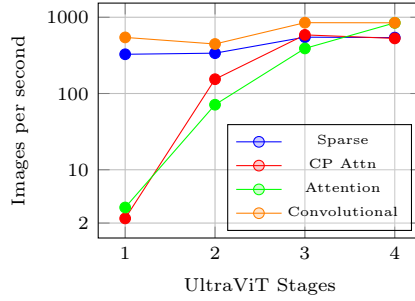
\begin{wrapfigure}
    [13]{r}{0.48\textwidth}
    \centering
  \vspace{-1.5cm}
    \begin{tikzpicture}
      \begin{axis}[
        xlabel={UltraViT Stages},
        ylabel={Images per second},
        legend style={
            font=\tiny, 
            fill opacity=0.2, 
            draw opacity=1, 
            text opacity=1
        },
        grid=major,
        xtick={1,2,3,4},
        xticklabels={1,2,3,4},
        ymode=log,
        ytick={2,10,100,1000},
        yticklabels={2,10,100,1000},
        legend pos=south east,
        font=\scriptsize,
        width=\linewidth,
        height=0.8\linewidth,
      ]
        \addplot[ blue, mark=*, table/skip first n=1 ] table [x index=0, y index=1,
        col sep=comma] {csvs/blocks_speed_balanced.csv}; \addlegendentry{Sparse}
        \addplot[ red, mark=*, table/skip first n=1 ] table [x index=0, y index=2,
        col sep=comma] {csvs/blocks_speed_balanced.csv}; \addlegendentry{CP Attn}
        \addplot[ green, mark=*, table/skip first n=1 ] table [x index=0, y
        index=3, col sep=comma] {csvs/blocks_speed_balanced.csv}; \addlegendentry{Attention}
        \addplot[ orange, mark=*, table/skip first n=1 ] table [x index=0, y index=4,
        col sep=comma] {csvs/blocks_speed_balanced.csv}; \addlegendentry{Convolutional}
      \end{axis}
    \end{tikzpicture}
    \caption{Inferences per second for images with resolution 512x512 at each stage.}
    \label{fig:blocks}
  \end{wrapfigure}

  We postulate that the optimal mixer operation, at different stages of the
  network, varies as a ratio of accuracy-to-latency trade-off. Hence, to optimize
  the architecture for edge deployment, we employ a systematic block selection
  process tailored to each stage of the network. The computational dynamics of different
  spatial mixers: Convolutional, CP, Sparse, and Vanilla Attention,
  vary significantly depending on the feature resolution ($N$) and channel
  dimension ($C$). Therefore, instead of relying on theoretical FLOPs, complexity estimates, on measurements on proxy devices (GPUs/CPUs), we conduct
  rigorous on-device latency measurements for each block candidate across a comprehensive
  grid of configurations, varying the number of channels and tokens. To our knowledge,
  this is the first work to perform such a systematic block selection process
  for mobile vision encoders for LVLMs. The results are shown in Fig.~\ref{fig:blocks}.
  Based on these measurements, we then
  select a targeted subset of these configurations for each stage, train them using
  a standard SigLIP~\cite{zhai2023sigmoid} loss with multiple captions~\cite{bulat2024fff},
  and evaluate their feature representations in a zero-shot manner for image
  retrieval, classification, and post LLaVA-OV finetuning, as part of the LVLM. The configurations are subsequently ranked
  based on their combined on-device performance and accuracy. As the results in Table~\ref{tab:mixer_ablation} demonstrate, the optimal configuration is heterogeneous, with different mixers suitable at different stages. In particular, it employs highly efficient convolutional mixers during the high-resolution Stage 1, an interleaved combination of convolutional and sparse attentive mixers in Stage 2, channel-partitioned attention in Stage 3, and full single-head attention at the lowest resolution. The resulting, final architecture, is depicted in Fig.~\ref{fig:ultravit}, where $C_{1-4}$=[192, 512, 768, 1536] and $N_{1-4}$=[4, 7, 10, 2].
 
  \subsection{Overall architecture}

\textbf{Pyramidal Structure and Stem:} Our network adopts a hierarchical
  pyramidal architecture. The processing begins with
  a stem block that rapidly downsamples the input high-resolution image by a
  factor of 4. Following the stem, the
  architecture is divided into four distinct stages, progressively downsampling the
  spatial dimensions by a factor of two while expanding the channel capacity.
  The downsampling block bridging these stages follows~\cite{yun2024shvit} and consists of a 1x1 conv layers, a 3x3 strided DW conv layer, another 1x1 conv. layer, and a SE module~\cite{hu2018squeeze}.

      \textbf{Multi-scale Feature Aggregator:} To support
      diverse vision-language tasks, including dense prediction and spatial grounding, we incorporate a lightweight
      feature aggregator combining semantically
      rich features from later stages with high-resolution,
      detailed features from earlier stages. The features are first projected
  to a common spatial resolution via strided convolutions, concatenated along the channel dimension, and finally fused through a $1 \times 1$ convolution to produce the aggregated representation (see Fig.~\ref{fig:ultravit}, bottom).
  
\begin{table}[!ht]
\caption{Accuracy-performance tradeoffs for different model configurations. Zero-shot results for classification (Imagenet, top-1) and image retrieval (a suite of 5 OCR datasets, @R1). LLaVA finetuning results reported at patch resolution of 512x512px, as part of a 1.5B LVLM.}
\label{tab:mixer_ablation}
\centering
\small
\resizebox{\textwidth}{!}{
\begin{tabular}{@{}lccc|cc|cccccc@{}}
\toprule
Mixer per Stage 
& Params (M) 
& Flops (B) 
& Inf. 
& ImageNet 
& OCR 
& TextVQA 
& DocVQA 
& InfoVQA 
& RWQA 
& MME 
& Avg. \\
\midrule
\textbf{C, C+S, CP, A} & 139.4 & 143.9 & 148.9 & 67.8 & \underline{33.4} & \underline{59.1} & \underline{64.8} & 39.8 & \textbf{57.9} & \underline{71.1} & \textbf{58.5} \\
C, C, CP, A & 142.4 & 144.7 & 147.8 & \underline{68.0} & 33.2 & 58.2 & \textbf{65.4} & 39.5 & \underline{57.7} & 68.5 & \underline{57.9} \\
C, C+S, A, A & 144.8 & 154.7 & 126.7 & \textbf{68.3} & \textbf{33.6} & \textbf{59.4} & 63.9 & \textbf{40.7} & \textbf{57.9} & 70.6 & \textbf{58.5} \\
C, C, C, C & 143.4 & 141.8 & 171.0 & 67.2 & 31.4 & 57.1 & 62.4 & 39.4 & 56.1 & 67.2 & 56.4 \\
C, C, C, A & 143.4 & 147.2 & 160.7 & 67.8 & 32.6 & 58.1 & 63.7 & 38.5 & 56.5 & 67.6 & 56.9 \\
\bottomrule
\end{tabular}
}
\end{table}

\section{Efficient Vision Encoder pre-training for LVLMs}

  Going beyond standard contrastive (CLIP or SigLIP) training, in this section, we present a new sample-efficient pre-training strategy for vision encoders tailored for LVLMs. The training strategy is based on two ideas: dense distillation and generative pre-training. During the first stage, we distil the knowledge of a large teacher model into our student model using a modified dense distillation. During the second stage,
  we use generative pre-training against a pretrained, frozen LLM to further improve the performance of our student model. 
  
  \subsection{Dense distillation}
  \label{sec:dense_distill}

  To effectively train the model, we propose to transfer the rich spatial and semantic knowledge of a large-scale
  pretrained teacher model into our compact architecture using a tailored
  dense distillation strategy. A primary challenge in cross-architecture distillation arises from the mismatch in resolution and feature dimensionality between the teacher and student.

  Rather than naively downsampling the teacher's high-resolution feature maps to match the student, a process that incurs a loss of fine-grained
  spatial information crucial for dense prediction tasks, we primarily preserve the spatial detail of the features via structural re-alignment. Specifically, denote the teacher features as $F_{T}\in \mathbb{R}^{H_T \times W_T \times C_T}$ and the
  student features as $F_{S}\in \mathbb{R}^{H_S \times W_S \times C_S}$. When the spatial resolutions are compatible ($H_{T}/ H_{S} = r$ is an integer), we apply Pixel Shuffle \cite{shi2016real} to rearrange the student’s tokens by reshaping local spatial blocks into the channel dimension. For an $r \times r$ spatial block, the reshaped student feature becomes $\hat{F}_{S}\in \mathbb{R}^{H_S \cdot r \times W_S \cdot r \times \frac{C_{S}}{r^{2}}}$. In cases where exact spatial alignment cannot be achieved, e.g., due to non-integer scaling factors or channel constraints, we apply a resolution-aligned downsampling of the teacher features to the closest compatible resolution. This preserves the overall alignment structure while avoiding excessive distortion of the teacher’s representations.

  Following this spatial-to-channel rearrangement, we apply a dual-objective
  loss to guide the student. First, we compute a dense cosine similarity loss applied patch-by-patch between the student features and the re-aligned teacher
  features, forcing the student to mimic the teacher's localized representational distribution:
  \begin{equation}
    \mathcal{L}_{dense}= 1 - \frac{1}{r^{2}H_{S}W_{S}}\sum_{i=1}^{H_T}\sum_{j=1}^{W_T}
    \frac{\hat{F}_{S}(i,j) \cdot F_{T}(i,j)}{\|\hat{F}_{S}(i,j)\| \|F_{T}(i,j)\|}.
  \end{equation}
  Second, to ensure that the student's global semantic understanding is preserved, we pass both the student's and teacher's features through the teacher's frozen attention pooling block, denoted as $\mathcal{P}_{T}(\cdot)$. We then apply a secondary cosine similarity loss on these aggregated, global representation tokens:
  \begin{equation}
    \mathcal{L}_{global}= 1 - \frac{\mathcal{P}_{T}(\hat{F}_{S}) \cdot
    \mathcal{P}_{T}(F_{T})}{\|\mathcal{P}_{T}(\hat{F}_{S})\| \|\mathcal{P}_{T}(F_{T})\|}
  \end{equation}

  The final dense distillation objective is simply a weighted combination of these
  two alignment losses: $\mathcal{L}_{distill}= 0.5 \mathcal{L}_{dense} + 0.5 \mathcal{L}_{global}$.

  \subsection{Generative pre-training}
  \label{sec:generative_pretraining}

  Generative pre-training of the vision encoder utilizing an autoregressive next-token prediction task has historically received limited attention compared to contrastive methods. This limited adoption stems from two critical drawbacks: the substantial computational cost inherent to training autoregressive decoders, and their fundamental unsuitability for straightforward zero-shot
  deployment, the very capability that popularized models like CLIP.
  Consequently, while recent works such as GIT~\cite{wang2022git}, CapPa~\cite{tschannen2023image},
  and the captioning extension in SigLIP~2~\cite{tschannen2025siglip} have demonstrated the conceptual viability of generative representation learning, these methods have not yet shown notable gains over contrastive methods for LVLMs and exhibit fundamental limitations.

  First, while SigLIP~2 incorporates a generative captioning objective alongside contrastive learning, empirical evidence shows that post-LLaVA fine-tuning, its performance is highly similar to the original SigLIP model~\cite{cocchi2025llava}. This suggests that their specific generative formulation does not yield representations that are uniquely advantageous for downstream LVLM tasks. Second, prior methods typically employ extremely small textual decoders during the pre-training phase. This design choice inadvertently shifts the linguistic and syntactic modeling burden onto the vision encoder, forcing it to allocate its limited representational capacity to non-visual tasks. For mobile-centric, parameter-constrained vision architectures, this waste of representational capacity is highly detrimental to dense spatial grounding.

  To address these shortcomings, we propose a tailored generative pre-training strategy. Crucially, the target Language Model (LLM) acting as the decoder must be explicitly pre-aligned to vision. Once alignment is achieved, we freeze the LLM. By utilizing a strong, pre-aligned LLM as the frozen decoder, we relieve the vision encoder of the linguistic modeling burden, ensuring its capacity is dedicated entirely to extracting rich visual features. Furthermore, to optimize training efficiency and accelerate convergence without compromising semantic alignment, we dynamically alternate the frozen decoder between a 0.5B and a 1.5B parameter model during the generative pre-training phase. We note that fully utilizing smaller models shifts the burden of language modeling again to the vision encoder. This model-switching strategy allows us to achieve deep alignment with high-capacity
  LLMs, while significantly reducing the overall computational cost of the
  pre-training pipeline.

  \section{Experiments}
  \label{sec:experiments}

  \subsection{Experimental Setup}
  Our training pipeline consists of three distinct phases: dense pre-training, generative
  pre-training, and supervised fine-tuning.

  \noindent
  \textbf{1. Dense Distilation Pre-training:} The model is trained on a randomly sampled subset
  of 150M examples from DataComp-1B~\cite{vasu2025fastvlm}. We optimize the
  dense objective of Sec.~\ref{sec:dense_distill} for 25 epochs using a cosine learning rate schedule, a
  peak learning rate of $1 \times 10^{-4}$, a weight decay of 0.1, and a batch size of 32k.
  
  \noindent
  \textbf{2. Generative Pre-training:} Initialized from the dense pre-trained weights,
  the model is trained for 1 epoch on the 85M data mix from~\cite{an2025llava}. We
  employ a cosine schedule with a learning rate of $1 \times 10^{-5}$, no weight decay, and a batch size of 192. The generative signal is provided through a frozen
  Qwen2 LLM, which was previously vision-aligned using 4M LLaVA-OV samples. The resulting
  model represents our final vision encoder, referred to throughout this paper as \textbf{UltraViT}.

  \noindent
  \textbf{3. Supervised Fine-tuning:} The final phase evaluates UltraViT within an LVLM. We substitute the original vision
  encoder in the LLaVA-OV architecture with UltraViT, resulting in an LVLM 
  coined \textbf{UltraVLM}. To guarantee direct and fair comparisons, UltraVLM strictly
  follows the 3-stage training recipe of LLaVA-OV~\cite{li2024llava-ov},
  utilizing a total of 7.1M samples: 4M OCR and captioning examples for Step 2, and
  3.1M instructionally-guided examples for Step 3.

  All models were trained on 32 H100 using PyTorch~\cite {paszke2019pytorch}
  and deepspeed~\cite{rasley2020deepspeed}.

  \noindent
  \textbf{On-device Benchmarking Protocol:} All mobile NPU inference latencies and
  throughput measurements reported in this work, both during our block-level
  latency search (Sec.~\ref{sec:spatial_mixers}) and full-model evaluation, were
  fully measured on physical hardware. Specifically, models were compiled using the
  Qualcomm Neural Processing SDK (QNN), fully quantized to INT8 precision, and benchmarked
  natively on a Samsung Galaxy S25 Ultra smartphone.

  \subsection{Comparison with state-of-the-art}

  We primarily compare UltraVLM with the current state-of-the-art approach for
  efficient vision encoders in LVLMs, FastVLM~\cite{vasu2025fastvlm}, evaluating
  it on a large suite of benchmarks including GQA~\cite{hudson2019gqa}, SQA~\cite{lu2022learn}, TextVQA~\cite{singh2019towards}, POPE~\cite{li2023evaluating}, DocVQA~\cite{mathew2021docvqa}, InfoVQA~\cite{mathew2022infographicvqa},
  RealWorldQA, MME~\cite{zhang2021mme}, MMMU~\cite{yue2024mmmu}, ChartQA~\cite{masry2022chartqa}, and MMSTAR~\cite{chen2024we}. To ensure a fair comparison, we
  retrain FastVLM using the same LLaVA-OV training recipe as UltraVLM, starting from their vision encoder, prior to their LVLMs training. For
  completeness, we also include the results reported in the original FastVLM
  paper under the data setting most comparable to ours. Additionally, to place
  our results in the context of the broader LVLM landscape, we include the performance
  of LLaVA-OV~\cite{li2024llava-ov}, QwenVL-2, QwenVL-2.5, and QwenVL-3. These are evaluated under two settings: native full
  resolution (an unconstrained number of vision tokens) and, following ~\cite{vasu2025fastvlm}, restricted to a matching number of tokens per image. 

    \begin{table}[!ht]
    \caption{Comparison with state-of-the-art LVLMs.
    {\color{gray}Grayedout} entries denote models evaluated at full resolution,
    using an unrestricted number of tokens. All other entries are evaluated under
    the same number of tokens per image. * denotes results taken from ~\cite{vasu2025fastvlm}.
    Re-trained models are trained under the same LLaVA-OV training recipe. Vis. inf. - denotes the on-device inference speed of the corresponding
    vision encoder at a reference resolution of 512x512px. Note that our
    approach (1) outperforms FastVLM on most benchmarks, (2) while having a much
    faster vision encoder (see also Sec.~\ref{ssec:efficiency}).}
    \label{tab:sota_comparison}
    \centering
    \resizebox{\linewidth}{!}{%
    \begin{tabular}{lllc|ccccccccccc}
      \toprule Model                                                 & \rot{\shortstack[l]{Vis \\ Arch.}}    & \rot{Vis inf./sec}  & \rot{\shortstack[l]{Data \\ amount}} & \rot{GQA}         & \rot{SQA}         & \rot{TextVQA}     & \rot{POPE}        & \rot{DocVQA}      & \rot{InfoVQA}     & \rot{RealWQA} & \rot{MME}         & \rot{MMMU}        & \rot{ChartQA}     & \rot{MMSTAR}      \\
      \midrule \multicolumn{15}{c}{\textit{Pre-trained baselines}}    \\
      \midrule \color{gray} LLaVA-OV~\cite{li2024llava-ov}           & \color{gray} SigLIP-400M  & \color{gray} 7.6 & \color{gray} 7.1M & \color{gray} 58.6 & \color{gray} 75.6 & \color{gray} 69.9 & \color{gray} 88.1 & \color{gray} 76.6 & \color{gray} 46.7 & \color{gray} 57.3 & \color{gray} 63.7 & \color{gray} 38.6    & \color{gray} 64.2 & \color{gray} 40.5 \\
      \color{gray} QwenVL-2~\cite{wang2024qwen2vl} \color{gray}      & \color{gray} QwenVL-2     & \color{gray} 3.7 & \color{gray} -    & \color{gray} 60.1 & \color{gray} 77.8 & \color{gray} 79.4 & \color{gray} 87.7 & \color{gray} 89.5 & \color{gray} 63.9 & \color{gray} 61.2 & \color{gray} 75.1 & \color{gray} 38.1 & \color{gray} 75.0 & \color{gray} 43.5 \\
      \color{gray} QwenVL-2.5 ~\cite{bai2025qwen25vltechnicalreport} & \color{gray} QwenVL-2.5   & \color{gray} 6.9 & \color{gray} -    & \color{gray} 60.0 & \color{gray} 80.7 & \color{gray} 78.6 & \color{gray} 87.3 & \color{gray} 92.4 & \color{gray} 75.0 & \color{gray} 59.1 & \color{gray} 75.8 & \color{gray} 39.6 & \color{gray} 83.7 & \color{gray} 56.2 \\
      \color{gray} QwenVL-3~\cite{bai2025qwen3vl}                    & \color{gray} SigLIP2-400M & \color{gray} 7.6 & \color{gray} -    & \color{gray} 59.4 & \color{gray} 86.3 & \color{gray} 79.2 & \color{gray} 89.4 & \color{gray} 92.7 & \color{gray} 72.4 & \color{gray} 63.5 & \color{gray} 74.6 & \color{gray} 46.9 & \color{gray} 79.8 & \color{gray} 43.1 \\
      \midrule QwenVL-2~\cite{wang2024qwen2vl}                       & QwenVL-2                  & 3.7              & -                 & 59.3              & 77.3              & 66.8              & 86.6              & 64.5              & 30.7              & 51.2              & 72.1              & 38.1              & 49.4              & 43.0              \\
      QwenVL-2.5 ~\cite{bai2025qwen25vltechnicalreport}              & QwenVL-2.5                & 6.9              & -                 & 60.5              & 80.3              & 66.5              & 86.8              & 61.8              & 35.0              & 58.8              & 77.5              & 39.8              & 71.5              & 55.0              \\
      QwenVL-3~\cite{bai2025qwen3vl}                                 & SigLIP2-400M              & 7.6              & -                 & 59.5              & 86.1              & 68.8              & 89.5              & 67.8              & 34.6              & 61.3              & 73.7              & 47.1              & 73.9              & 43.1              \\
      \midrule

FastVLM~\cite{vasu2025fastvlm}*                      & FastViTHD                 & 91.4              & 16.1M             & 64.2              & 74.8              & 66.0              & 88.0              & 67.7              & -                 & -                 & -                 & 33.1              & -                 & -                 \\
      \midrule \multicolumn{15}{c}{\textit{Re-trained models}}        \\
      \midrule                                                       
      FastVLM~\cite{vasu2025fastvlm}                                 & FastViTHD                 & 91.4              & 7.1M              & 59.9              & 80.0              & 62.9              & \textbf{87.3}     & 66.7              & 43.5              & \textbf{60.1}     & 69.9              & \textbf{41.1}     & 66.0              & 44.1              \\
      UltraVLM                                                       & UltraViT                  & 148.9              & 7.1M              & \textbf{60.5}     & \textbf{82.9}     & \textbf{68.9}     & 87.1              & \textbf{71.8}     & \textbf{47.9}     & 59.5              & \textbf{70.0}     & 38.1              & \textbf{72.2}     & \textbf{49.8}     \\
      \bottomrule
    \end{tabular}
    }
  \end{table}
  
  As Table~\ref{tab:sota_comparison} shows, UltraVLM outperforms FastVLM, across most benchmarks, achieving
  particularly large gains on challenging tasks such as TextVQA (+6.0\%), DocVQA (+5.1\%),
  and ChartQA (+6.2\%), all while employing a much faster vision encoder (see also Sec.~\ref{ssec:efficiency}).
  It also outperforms the original FastVLM despite FastVLM being trained with more than twice
  the amount of supervision during the LVLM fine-tuning phase (+4.1\% on DocVQA,
  +2.9\% on TextVQA, etc.). Finally, when the QwenVL-2/2.5/3 series models are
  restricted to the same token budget (256 tokens) as ours, our UltraVLM surpasses them on the majority of benchmarks. This highlights UltraVLM's
  efficiency, especially considering that the QwenVL models utilize much larger and
  slower vision encoders, more modern LLM backbones, and are trained on significantly
  larger pools of proprietary private data. Though we report the unconstrained
  full-resolution results for these models for completeness, operating them without
  token constraints yields LVLMs that are more than an order of magnitude slower
  than UltraVLM.

    \begin{table}[!ht]
    \centering
    \caption{Comparison with token reduction methods. Unlike post-hoc pruning approaches
    that hurt accuracy and leave vision encoder latency unchanged, UltraVLM's
    pyramidal design natively achieves a massive $16.0\times$ token reduction.}
    \label{tab:comparison_with_token_reduction} \resizebox{\linewidth}{!}{%
    \begin{tabular}{l|cccccccccc|cc}
      \toprule \textbf{Method}                       & \rot{GQA}     & \rot{SQA}     & \rot{TextVQA} & \rot{POPE} & \rot{DocVQA}  & \rot{InfoVQA} & \rot{RealWQA} & \rot{MME}     & \rot{ChartQA} & \rot{MMSTAR}  & \rot{{\begin{tabular}[c]{@{}c@{}}Avg. num. \\ tokens \\ reduction\end{tabular}}} & \rot{{\begin{tabular}[c]{@{}c@{}}Vis Enc \\ Speedup\end{tabular}}} \\
      \midrule LLaVA-OV \cite{li2024llava-ov}        & 58.6          & 75.6          & \textbf{69.9}          & \textbf{88.1}       & \textbf{76.6}          & 46.7          & 57.3              & 63.7          & 64.2          & 40.5          & $1.0\times$                                                                             & $1.0\times$                                                               \\
      \midrule 
      VisionZip \cite{yang2025visionzip} & 57.2          & 76.2          & 62.0          & 87.0       & 55.1          & 30.8          & 54.6              & 63.9          & 50.0          & 38.1          & $5.7\times$                                                                             & $1.0\times$                                                               \\
      PyramidDrop \cite{xing2025pyramiddrop}         & 55.9          & 76.0          & 60.3          & 87.2       & 54.5          & 34.2          & 56.7              & 63.7          & 47.9          & 37.5          & $4.6\times$                                                                             & $1.0\times$                                                               \\
      VisPruner \cite{zhang2025beyond}               & 53.4          & 76.0          & 54.8          & 83.3       & 51.2          & 28.4          & 52.9              & 62.3          & 40.0          & 38.4          & $5.7\times$                                                                             & $1.0\times$                                                               \\
      HiRED \cite{arif2025hired}                     & 56.3          & 76.0          & 59.8          & 85.9       & 52.3          & 27.6          & 54.8              & 63.2          & 44.1          & 39.3          & $5.0\times$                                                                             & $1.0\times$                                                               \\
      \midrule \textbf{UltraVLM}                     & \textbf{60.5} & \textbf{82.9} & 68.9 & 87.1       & 71.8 & \textbf{47.9} & \textbf{59.5}     & \textbf{70.0} & \textbf{72.2} & \textbf{49.8} & $16.0\times$                                                                            & $19.1\times$                                                                        \\
      \bottomrule
    \end{tabular}%
    }
  \end{table}

  \subsection{Comparison with token reduction methods}

  In addition to utilizing a highly efficient vision encoder, UltraVLM organically
  achieves a significant $16.0\times$ reduction in the number of tokens without any external
  token reduction mechanisms. Herein, we compare UltraVLM with previously proposed state-of-the-art token reduction methods. As shown in Table~\ref{tab:comparison_with_token_reduction}, UltraVLM comfortably outperforms all existing reduction strategies by significant margins across every benchmark. Crucially, because our token reduction is an intrinsic property of the underlying architecture, UltraVLM
  is the only method that additionally delivers a significant inference speed-up for
  the vision encoder itself. Note that all methods use the same LLM and training data.

  \subsection{Inference and train-time efficiency}~\label{ssec:efficiency}

  \noindent \textbf{Inference efficiency:} \textit{Compared to the commonly adopted SigLIP or
  SigLIP-2 400M:} First,
  the newly proposed UltraViT vision encoder executes $19.1\times$ faster during inference. Secondly, our pyramidal architecture intrinsically outputs
  $16\times$ fewer visual tokens, directly shrinking the input sequence length and slashing the computational burden of the LLM proportionally.  
  \textit{Compared to FastVLM:} UltraViT operates nearly $1.7\times$ faster
  than FastViT at a standard $512 \times 512$px resolution.
  Furthermore, as illustrated in Fig.~\ref{fig:ultravit_vs_fastvit_speed}, this relative
  speed-up remains highly consistent across a wide range of input resolutions. As noted before, all models are benchmarked
  natively on a Samsung Galaxy S25 Ultra.

    \begin{figure}[!ht]
    \centering
    \begin{minipage}[c]{0.48\linewidth}
      \centering
      \begin{tikzpicture}
        \begin{axis}[
          xlabel={Input image resolution},
          ylabel={Inf./sec.},
          ymode=log,
          grid=major,
          log basis y=10,
          ytick={2,5,10,25,50,100,250,500},
          yticklabels={2,5,10,25,50,100,250,500},
          xtick={1,2,3,4,5,6},
          xticklabels={256,384,512,1024,1536,2048},
          legend pos=south west,
          font=\scriptsize,
          width=\linewidth,
          height=0.8\linewidth,
        ]
          \addplot[ red, mark=*, table/skip first n=1 ] table [x index=0, y
          index=1, col sep=comma] {csvs/full_model_res.csv}; \addlegendentry{FastViT}
          \addplot[ blue, mark=*, table/skip first n=1 ] table [x index=0, y index=2,
          col sep=comma] {csvs/full_model_res.csv}; \addlegendentry{UltraViT}
        \end{axis}
      \end{tikzpicture}
      \vspace{-0.4cm}
      \caption{Inferences per second for UltraViT vs FastViT for varying input image resolutions.}
      \label{fig:ultravit_vs_fastvit_speed}
    \end{minipage}\hfill
    \begin{minipage}[c]{0.48\linewidth}
      \makeatletter\def\@captype{table}\makeatother
      \centering
      \vspace{-0.5cm}
      \caption{Pre-training efficiency. UltraViT achieves state-of-the-art
      results while processing $3\times$ fewer total samples and $10\times$ fewer unique images than FastVLM.}
      \label{tab:pretraining_cost}
      \vspace{0.5em} 
      \resizebox{0.9\linewidth}{!}{%
      \begin{tabular}{@{}llrr@{}}
        \toprule \textbf{Model}            & \textbf{Stage} & \textbf{Data} & \textbf{Seen} \\
        \midrule FastVLM                   & Contrastive    & 2B            & 13B           \\
        \midrule \multirow{3}{*}{UltraViT} & Dense Distill. & 150M          & 4B            \\
                                           & Generative     & 85M           & 85M           \\
        \cmidrule{2-4}                     & \textbf{Total} & \textbf{235M} & \textbf{4.1B} \\
        \bottomrule
      \end{tabular}%
      }
    \end{minipage}
  \end{figure}

  \noindent
  \textbf{Pre-training efficiency:} While prior contrastive strategies
  demand massive data regimes to converge, our two-stage pre-training framework (dense distillation followed by generative supervision) achieves superior performance with a fraction of the samples. As detailed in Table~\ref{tab:pretraining_cost},
  compared with FastVLM, UltraViT achieves state-of-the-art results while seeing approximately $3\times$ fewer total training samples (4.1B
  vs.\ 13B) and processing nearly $10\times$ fewer unique images (235M vs.\ 2B) compared
  to the contrastive distillation employed by FastVLM. 

  \section{Ablation Studies}

  \noindent
  \textbf{Impact of the dense distillation:} Table~\ref{tab:ablations_combined}(a) compares our dense distillation with standard contrastive (SigLIP) training. 
  With UltraViT, dense distillation brings consistent gains, with larger improvements on fine-grained tasks such as DocVQA and InfoVQA. Further training with a larger teacher improves performance, indicating that our approach scales well with stronger teachers and longer training.
  Similarly, for FastViT, dense distillation consistently outperforms contrastive training across all benchmarks. Under the same training setup, UltraViT achieves comparable performance to FastViT while running significantly faster on-device.

\begin{table}[!ht]
\centering
\caption{Ablation studies on the proposed Dense distillation and Generative pre-training techniques.}
\label{tab:ablations_combined}
\setlength{\tabcolsep}{3pt}
\renewcommand{\arraystretch}{1.05}
\begin{minipage}{0.495\linewidth}
\centering
\textbf{(a) Dense Distillation}\\[4pt]
\resizebox{\linewidth}{!}{%
\begin{tabular}{llcccc}
\toprule
Model & Variant & DocVQA & InfoVQA & MMSTAR & SQA \\
\midrule
\multirow{3}{*}{UltraViT} & Contrastive & 58.9 & 36.4 & 42.3 & 75.8 \\
& Dense Distil. & 63.9 & 39.5 & 44.6 & 76.9 \\
& + Large Teacher & 67.5 & 42.8 & 44.4 & 78.5 \\
\midrule
\multirow{2}{*}{FastViT} & Contrastive & 60.7 & 39.1 & 44.4 & 74.0 \\
& Dense Distil. & 62.1 & 39.4 & 45.9 & 76.3 \\
\bottomrule
\end{tabular}
}
\end{minipage}
\hfill
\begin{minipage}{0.495\linewidth}
\centering
\mbox{\textbf{(b) Gen. Pre-training (UltraViT)}}\\[4pt]
\resizebox{\linewidth}{!}{%
\begin{tabular}{lcccc}
\toprule
Variant & DocVQA & InfoVQA & MMSTAR & SQA \\
\midrule
Dense Distil. (+ LT). & 67.5 & 42.8 & 44.4 & 78.5 \\
+ Gen (0.5B) & 70.0 & 44.7 & 47.3 & 81.5 \\
+ Gen (1.5B) & 72.0 & 46.8 & 49.1 & 83.1 \\
\midrule
+ Gen (Dyn.) & 71.8 & 47.9 & 49.8 & 82.9 \\
\bottomrule
\end{tabular}
}
\end{minipage}
\end{table}

  \noindent
  \textbf{Impact of generative pre-training:} Herein, we ablate the impact of the generative pre-training stage for the UltraViT model and the choice of the LLM decoder. As shown in Table~\ref{tab:ablations_combined}(b), further fine-tuning the vision
  encoder using a frozen LLM results in notable accuracy gains. When utilizing a small 0.5B LLM, performance improves, but to a smaller degree; the limited linguistic capacity of the small decoder forces the vision encoder to compensate for language modeling, bottlenecking visual representation quality. Utilizing a 1.5B LLM yields strong semantic alignment but incurs a larger computational pre-training overhead. Our proposed dynamic switching strategy, which alternates between the 0.5B and 1.5B decoders, matches or exceeds the performance of the pure 1.5B model while notably reducing the total training cost.

\section{Conclusion}

In this paper, we presented UltraViT, a latency-optimized vision encoder explicitly designed for deploying Large Vision-Language Models (LVLMs) on edge devices. Our contributions are two-fold: First, we introduce a novel heterogeneous architecture guided by real on-device latency measurements. Then, we propose a novel two-stage pre-training strategy that combines dense feature distillation with generative LLM supervision.  Extensive empirical evaluations demonstrate that our holistic co-design establishes a new state-of-the-art for efficient LVLM encoding. UltraViT significantly outperforms existing on-device baselines, such as FastVLM, across diverse multimodal benchmarks while operating at nearly 2x the speed on mobile hardware.

  \clearpage 

  \bibliographystyle{splncs04}
  \bibliography{main}

  \clearpage

\appendix

  \section{Additional Implementation Details}
  For the majority of our dense distillation experiments, we use SigLIP2-B16 NaFlex\footnote{\url{https://huggingface.co/google/siglip2-base-patch16-naflex}}
  as the teacher model. For the \texttt{+ Large Teacher} setting reported in
  Table~5(a), we resume training from the dense-distilled checkpoint and replace
  the teacher with a larger SigLIP2-400M model\footnote{\url{https://huggingface.co/google/siglip2-so400m-patch16-512}}.

\begin{figure}
    \centering
    \includegraphics[width=1\linewidth]{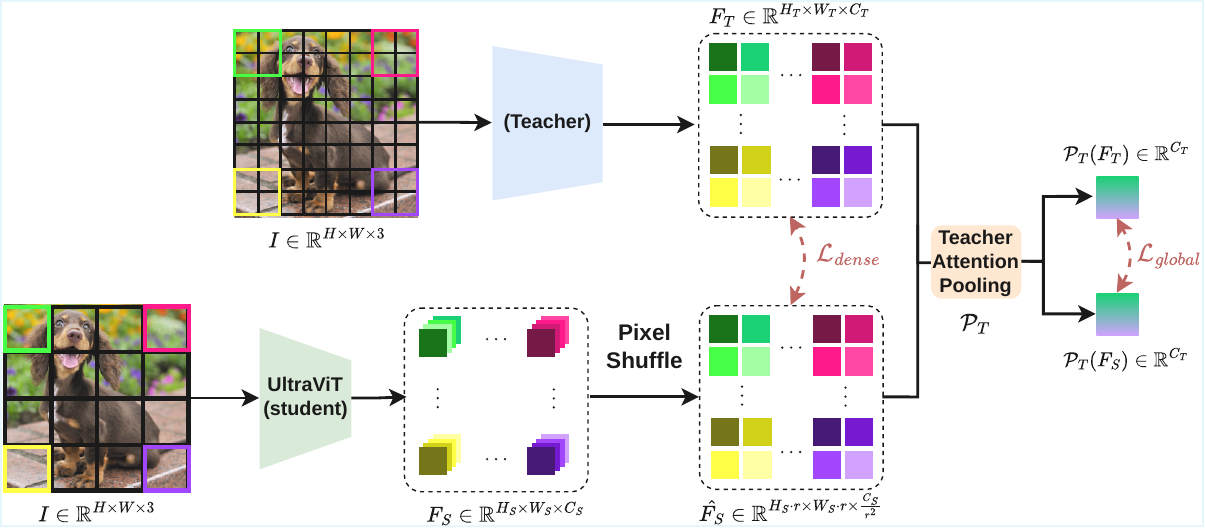}
    \caption{Overview of the proposed dense distillation strategy. The teacher and
    UltraViT (student) extract multi-scale features from the same input. To
    address resolution mismatch, student features are re-aligned via Pixel Shuffle
    before applying a dense patch-wise alignment loss ($\mathcal{L}_{dense}$).
    Both re-aligned student and teacher features are further processed through
    the teacher’s attention pooling module $\mathcal{P}_{T}$ to enforce global
    semantic consistency ($\mathcal{L}_{global}$).}
    \label{fig:dense}
  \end{figure}

  Fig.~\ref{fig:dense} illustrates the overall pipeline of the proposed dense
  distillation strategy.

  \section{Additional Experiments}

  \subsection{Extended OCR retrieval results}
  During the block selection process (Section~3.2), we compared different token
  mixer configurations in a zero-shot setting on multiple benchmarks, including image
  classification, OCR-based retrieval, and post LLaVA-OV fine-tuning.

  Due to space constraints, the main paper only reports the averaged OCR retrieval
  score. Here, we detail the five OCR datasets used. These include TextCaps~\cite{sidorov2020textcaps}
  and four VisRAG-Ret~\cite{yuvisrag} datasets: VisRAG-Ret-SlideVQA\footnote{https://huggingface.co/datasets/openbmb/VisRAG-Ret-Test-SlideVQA},
  VisRAG-Ret-InfoVQA\footnote{https://huggingface.co/datasets/openbmb/VisRAG-Ret-Test-InfoVQA},
  VisRAG-Ret-ChartQA\footnote{https://huggingface.co/datasets/openbmb/VisRAG-Ret-Test-ChartQA},
  and VisRAG-Ret-ArxivQA\footnote{https://huggingface.co/datasets/openbmb/VisRAG-Ret-Test-ArxivQA}.

  Table~\ref{tab:mixer_ocr} extends Table~1 from the main paper and reports the
  recall@1 performance on each individual OCR retrieval dataset, together with their
  average (AVG). While the configuration (C, C+S, A, A) achieves the highest average
performance across the five OCR benchmarks, the selected configuration (C, C+S, CP, A)
remains competitive, with only a marginal drop in accuracy, while offering better
efficiency.

\begin{table}[!ht]
\caption{Zero-shot OCR retrieval performance (R@1) for different mixer
    configurations across five datasets: TCAP (TextCaps), SVQA (SlideVQA), IVQA (InfoVQA),
    CHQA (ChartQA), and ARXQ (ArxivQA). AVG denotes the mean recall across all
    five datasets.}
\label{tab:mixer_ocr}
\centering
\small \resizebox{\textwidth}{!}{
\begin{tabular}{@{}lSSS|cccccc}
\toprule
{Mixer per Stage} & {Params (M)} & {Flops (B)} & {Inf.} & {TCAP} & {SVQA} & {IVQA} & {CHQA} & {ARXQ} & \textbf{AVG} \\
\midrule
\textbf{C, C+S, CP, A} & 139.4 & 143.9 & 148.9 & \textbf{66.6} & 40.3 & \underline{25.6} & \textbf{22.2} & 12.1 & \underline{33.4} \\
C, C, CP, A & 142.4 & 144.7 & 147.8 & \underline{66.3} & 41.0 & 24.1 & \underline{20.6} & \textbf{13.7} & 33.1 \\
C, C+S, A, A & 144.8 & 154.7 & 126.7 & \textbf{66.6} & \underline{41.5} & \textbf{27.7} & 19.0 & 13.1 & \textbf{33.6} \\
C, C, C, C & 143.4 & 141.8 & 171.0 & 66.0 & 37.8 & 24.7 & 17.5 & 11.2 & 31.4 \\
C, C, C, A & 143.4 & 147.2 & 160.7 & 65.8 & \textbf{43.2} & 25.1 & 15.9 & \underline{13.2} & 32.6 \\
\bottomrule
\end{tabular}
}
\end{table}

  \subsection{Effect of different teachers for dense pretraining}

  In this section, we investigate the impact of utilizing different teacher models
  during the dense pretraining phase for UltraViT. We evaluate four variants
  using different teacher targets: SigLIP2-NF (naflex)~\cite{tschannen2025siglip},
  Perception Encoder (PE)~\cite{bolya2025perception}, SigLIP2~\cite{tschannen2025siglip},
  and TULIP~\cite{tang2025tulip}. The detailed results across eleven distinct vision-language
  benchmarks are presented in Table~\ref{tab:teacher_comparison}.

  As shown in the table, employing strong, feature-rich vision models as teachers,
  such as SigLIP2 and TULIP, yields the best overall performance. Both SigLIP2
  and TULIP consistently outperform the other variants across the majority of
  tasks. However, overall, our approach is robust to the choice of teacher.

  \begin{table}[!ht]
    \caption{UltraVLM performance under different teachers during the dense
    pretraining phase.}
    \label{tab:teacher_comparison}
    \centering
    \resizebox{\linewidth}{!}{%
    \begin{tabular}{l|l|lcccccccccc}
      \toprule Model & Teacher    & \rot{GQA} & \rot{SQA} & \rot{TextVQA} & \rot{POPE} & \rot{DocVQA} & \rot{InfoVQA} & \rot{RealWQA} & \rot{MME} & \rot{MMMU} & \rot{ChartQA} & \rot{MMSTAR} \\
      \midrule 
      UltraVLM       & SigLIP2-NF & 58.5      & 76.9      & 57.1          & 86.3       & 63.9         & 39.5          & 53.1          & 68.9      & 33.1       & 62.8          & 44.6         \\
      UltraVLM       & PE         & 58.8      & 77.3      & 57.3          & 86.3       & 63.6         & 38.1          & 55.3          & 65.8      & 32.0       & 63.8          & 43.5         \\
      UltraVLM       & SigLIP2    & 59.5      & 78.8      & 59.4          & 86.7       & 65.3         & 40.3          & 57.3          & 69.4      & 33.9       & 64.5          & 43.6         \\
      UltraVLM       & TULIP      & 59.5      & 78.6      & 59.6          & 86.7       & 65.1         & 40.9          & 56.3          & 69.1      & 33.6       & 64.1          & 43.6         \\
      \bottomrule
    \end{tabular}%
    }
  \end{table}

\subsection{Efficient VLM comparisons}

We expand on our experiments presented in Table 4 of the paper comparisons with VLMs designed with the goal of being efficient. We present these results in Table~\ref{tab:effvlm}. There, in addition to performance measurements on two popular benchmarks, we report the inference times for those VLMs' vision encoders for a fixed input resolution, the Time To First Token (TTFT) for those VLMs following their inference recipe (i.e. using the native resolution and patch strategy recommended by each model), and the peak memory consumption of each vision encoder during inference on-device. These resoluts demonstrate that, even compared to other highly efficient VLMs, our approach is significantly faster, showcasing the advantages of designing the vision encoder architecture specifically for on-device deployment.

\begin{table}[t]
\centering
\caption{Vision-only and end-to-end TTFT; peak vision-only RAM.}
\begin{tabular}{lccccc}
\toprule
                 & Vis.\ (ms) & TTFT (ms) & Peak    & Text & ChartQA \\
                 & 512$^2$/1024$^2$ & 512$^2$/1024$^2$ & RAM(MB) & VQA  &       \\
\midrule
LLaVA-OV(SigLIP)        & 127.8/2,200.0 & 483/2555 & 300/349 & 69.9 & 64.2 \\
SmolVLM            & 36.6/690.1  & 100.6/755.1  & 77/92 & 60.5 & 62.8 \\
Florence-VL          & 305.2/11,837.8     & 569/12101  & 422/584  & 69.1 & 70.7 \\
FastVLM\,$\dagger$ (re-tr.)   & 10.9/65.9  & 98/154 & 66/91 & 62.9 & 66.0 \\
\textbf{UltraVLM (ours)}      & 6.6/41.0 & 94/129 & 60/75 & 68.9 & 72.2 \\
\bottomrule
\end{tabular}
\label{tab:effvlm}
\end{table}

\subsection{On-device measurements methodology}

As mentioned in the paper, on-device inference latency measurements were performed on a Samsung Galaxy S25 Ultra smartphone, with models compiled using the
  Qualcomm Neural Processing SDK. More specifically, to mitigate randomness in our measurements, for each model we ran 7 rounds of 100 inference passes.
  We subsequently ignore the fastest and slowest rounds, and report the average inference speed of the 5 remaining rounds. 
  To decrease the impact of the device overheating and interference from other on-device factors, we include wait times of 10s between rounds, and reboot the device between model measurements.
  
For block-level measurements (Figure 2) we stack 10 blocks of a given type and perform measurements for various stages of our model. For each stage, the block's width follows UltraViT's configuration as described in Section 3, and the input tensor corresponds to what those blocks would be processing in the corresponding stage of the model for an input image of resolution 512×512. E.g. for Stage 1 measurements, the blocks have width $C=192$ and the input tensor to the block stack has shape $(192, 64, 64)$. For Figure 3, we vary the number of blocks per block type so that, across types, we have approximately the same number of parameters.

\subsection{GPU and CPU performance}

In this section we contrast UltraViT's performance with FastViT on CPUs and GPUs. These measurements are made for input images of resolution 512x512 on a Nvidia RTX4090 GPU and a Inter i7-14700K CPU.
For GPU measurements we set the batch size to 96 (close to the limit of our GPU's capacity). For CPU measurements, we convert our model to onnx format and run single-image inferences. In both cases, we report  the inference time per sample processed. Similarly to the on-device measurements, we conduct 7 rounds of measurements with 20 inferences per round, and average the latencies excluding the fastest and slowest rounds for robustness.
As seen in~\ref{tab:cpugpu}, UltraViT, despite being designed specifically for optimal on-device speed, outperforms FastViT in both CPUs and GPUs, highlighting the efficiency of our proposed architecture.

  \begin{table}[!ht]
    \caption{Inference times on CPU and GPU for UltraViT and FastViT.}
    \label{tab:cpugpu}
    \centering
    \small
    {\setlength{\tabcolsep}{20pt}
    \begin{tabular}{lcc}
      \toprule Model & CPU (ms/sample)    & GPU (ms/sample)  \\
      \midrule
      FastViT & 118.0 & 2.5 \\
      UltraViT & \textbf{72.7} & \textbf{2.2} \\
      \bottomrule
    \end{tabular}%
    }
  \end{table}

\end{document}